\documentclass[letterpaper, 10 pt, conference]{ieeeconf}  % Comment this line out if you need a4paper

\IEEEoverridecommandlockouts                              % This command is only needed if 
\title{\LARGE \bf
GPO: Growing Policy Optimization for Legged Robot \\
    Locomotion and Whole-Body Control
}

\author{
Shuhao Liao$^{1,2}$, Peizhuo Li$^{2}$, Xinrong Yang$^{2}$, Linnan Chang$^{2}$, Zhaoxin Fan$^{1}$, Qing Wang$^{1}$,\\ Lei Shi$^{3}$, Yuhong Cao$^{2}$, Wenjun Wu$^{1}$, Guillaume Sartoretti$^{2}$ 
\thanks{$^{1}$Beihang University, China}
\thanks{$^{2}$Department of Mechanical Engineering, National University of Singapore, Singapore}
\thanks{$^{3}$Henan Univeristy, China}
}

\usepackage{amssymb}
\usepackage{amsmath}
\usepackage{graphicx}
\usepackage{multirow}
\usepackage{booktabs}
\usepackage{tabularx}
\usepackage{subcaption}
\usepackage{cite}
\usepackage{tcolorbox}
\usepackage[ruled]{algorithm2e}

\begin{document}

\maketitle
  
\thispagestyle{empty}
\pagestyle{empty}

%%%%%%%%%%%%%%%%%%%%%%%%%%%%%%%%%%%%%%%%%%%%%%%%%%%%%%%%%%%%%%%%%%%%%%%%%%%%%%%%
%%%%%%%%%%%%%%%%%%%%%%%%%%%%%%%%%%%%%%%%%%%%%%%%%%%%%%%%%%%%%%%%%%%%%%%%%%%%%%%%

\begin{abstract}

Training reinforcement learning (RL) policies for legged robots remains challenging due to high-dimensional continuous actions, hardware constraints, and limited exploration. 
Existing methods for locomotion and whole-body control work well for position-based control with environment-specific heuristics (e.g., reward shaping, curriculum design, and manual initialization), but are less effective for torque-based control, where sufficiently exploring the action space and obtaining informative gradient signals for training is significantly more difficult.
We introduce Growing Policy Optimization (GPO), a training framework that applies a time-varying action transformation to restrict the effective action space in the early stage, thereby encouraging more effective data collection and policy learning, and then progressively expands it to enhance exploration and achieve higher expected return.
We prove that this transformation preserves the PPO update rule and introduces only bounded, vanishing gradient distortion, thereby ensuring stable training.   
We evaluate GPO on both quadruped and hexapod robots, including zero-shot deployment of simulation-trained policies on hardware. 
Policies trained with GPO consistently achieve better performance.
These results suggest that GPO provides a general, environment-agnostic optimization framework for learning legged locomotion.

\end{abstract}

%%%%%%%%%%%%%%%%%%%%%%%%%%%%%%%%%%%%%%%%%%%%%%%%%%%%%%%%%%%%%%%%%%%%%%%%%%%%%%%%
%%%%%%%%%%%%%%%%%%%%%%%%%%%%%%%%%%%%%%%%%%%%%%%%%%%%%%%%%%%%%%%%%%%%%%%%%%%%%%%%

\section{INTRODUCTION}

Learning-based methods have advanced embodied artificial intelligence, enabling robots to acquire increasingly complex skills through interaction. This progress is particularly evident in legged robot locomotion and whole-body control, where control problems are inherently continuous, involve rich and intermittent contacts with the environment, and demand strong generalization across terrains, disturbances, and operating conditions. Among these approaches, RL has demonstrated strong potential by directly exploiting high-dimensional sensory feedback and complex nonlinear dynamics, resulting in agile and adaptive behaviors beyond the reach of conventional model-based controllers \cite{chen2024reinforcement,radosavovic2024real,li2025sata}. 
    Despite these advances, learning reliable policies for such systems remains challenging.
    Current learning-based methods often struggle to simultaneously achieve robust generalization and safe behavior, particularly during training and early deployment.
    
    \begin{figure}[!t]
        \includegraphics[width=\linewidth]{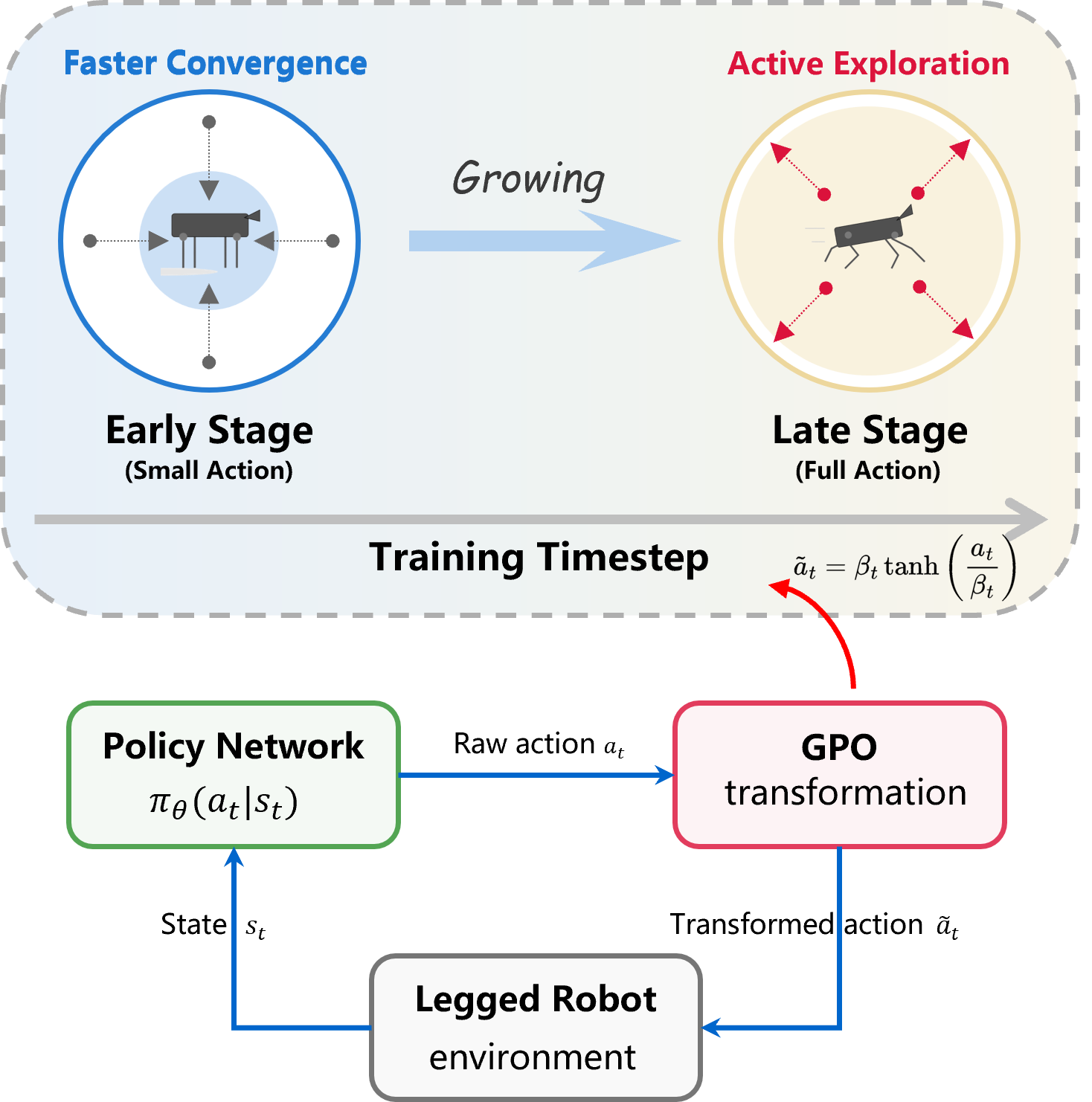}
        \setlength{\abovecaptionskip}{2pt}
        % \vspace{-0.3cm}
        \caption{Overview of GPO. GPO gradually expands the effective action space during training, enabling stable early optimization and improving asymptotic performance.}
        \label{fig:framework}
        % \vspace{-0.3cm}
    \end{figure}
    
    A central challenge is learning policies over high dimensional, continuous action spaces under strict torque, velocity, contact, and safety constraints.
    Moreover, real-world rollouts are costly, risky, and difficult to scale, limiting the data available for trial and error learning.
    To address these challenges, prior work has largely focused on modifying the training pipeline rather than the underlying optimization mechanism. 
    A common strategy is reward shaping, where task priors such as gait regularization, energy penalties, or contact heuristics are manually encoded to stabilize learning and bias policy search toward feasible locomotion patterns \cite{tsounis2020deepgait, yu2018learning, fu2021minimizing}. 
    Curriculum learning approaches gradually increase task difficulty by varying terrain complexity, initial states, or disturbance levels, improving performance on specific environments \cite{kobayashi2020reinforcement, berseth2018progressive, li2024learning, li2023robust}. 
    Other methods rely on expert demonstrations or supervised pre-training to warm-start RL in favorable regions of the policy manifold \cite{peng2018deepmimic, zhuang2023parkour, he2024learning}. 
    On the action side, practitioners use clipping, rescaling, or saturation of control commands to satisfy hardware limits and reduce destabilizing high variance actuation \cite{chen2023learning, kim2023torque, sombolestan2024adaptive}. 
    These works have enabled impressive results across diverse robotic platforms, suggesting that carefully managing exploration and action magnitudes is crucial for successful embodied RL \cite{lee2020learning, miki2022learning, rudin2022learning}.

    Despite these successes, most approaches modify the training pipeline but leave the optimization mechanism unchanged, and their effect on optimization properties is unclear. Reward shaping, curricula, and action clipping change the visited state-action distribution, but their impact on gradient variance, local convergence, and asymptotic performance is not well understood.
    Importantly, most widely used policy-gradient methods implicitly assume a fixed action space throughout training.   
    This design encourages wide exploration early in training, leading to high gradient variance and a rugged optimization landscape. As a result, the policy may fail to first learn reliable low-amplitude control before converging to an optimal policy.
    In practice, engineers often restrict the “effective” action space early (e.g., tighter torque limits or aggressive clipping) and relax it later \cite{lee2020learning,kim2023torque}. However, there is no principled framework that treats the evolution of the action space as part of the learning algorithm, nor a clear analysis of its effect on optimization and performance.
	
	In this work, we take a step toward addressing this gap by proposing GPO, a framework in which the effective action space is no longer fixed but gradually expands during training. GPO introduces a smooth, time-varying transformation that maps a fixed policy output to a growing range of executable actions. This design allows the agent to first learn reliable behaviors within a restricted action space, where the optimization is easier to navigate and gradient estimates are more stable, and then progressively unlock larger action space as training proceeds. 
    Crucially, this transformation is equivalent to PPO updates in structure: the underlying objective and clipping mechanism remain unchanged up to a change of variables, and we prove that the discrepancy between GPO gradients and standard PPO gradients is bounded. 
    Under mild assumptions, GPO achieves faster early-stage convergence and higher asymptotic return than PPO with a fixed action space.
    We validate these properties in simulation and on real robots under torque-based control, where actions are directly mapped to joint torques, leading to strong coupling and stringent stability requirements.
    Overall, GPO provides a general, environment-agnostic mechanism for improving training efficiency and final performance in RL-based robotic control.

%%%%%%%%%%%%%%%%%%%%%%%%%%%%%%%%%%%%%%%%%%%%%%%%%%%%%%%%%%%%%%%%%%%%%%%%%%%%%%%%
%%%%%%%%%%%%%%%%%%%%%%%%%%%%%%%%%%%%%%%%%%%%%%%%%%%%%%%%%%%%%%%%%%%%%%%%%%%%%%%%

\section{RELATED WORK}

Learning-based legged robot locomotion has advanced rapidly in recent years with the adoption of RL, enabling agile and dynamic behaviors~\cite{neunert2018whole,hu2019learning,ding2019real,kim2019highly,ren2021learning,bellegarda2022robust,grandia2023perceptive,bellegarda2024robust}.
    Many works focus on proprioceptive locomotion, where policies rely exclusively on onboard sensing data such as IMU and joint encoder measurements and are trained using actor-critic frameworks.
    To improve state evaluation and policy learning under partial observations, prior approaches incorporate privileged information during training~\cite{rudin2022learning}, historical observations via teacher-student distillation~\cite{lee2020learning,kumar2021rma,radosavovic2024real}, or direct state estimation from temporal windows~\cite{nahrendra2023dreamwaq}.
    These techniques have proven effective for reliable locomotion over diverse terrains and have established mature learning pipelines for legged robot control under limited sensing.
    Beyond state estimation, a wide range of methods aim to improve training efficiency in high-dimensional continuous control.
    Early approaches combine imitation learning with reinforcement learning to improve sampling efficiency, enabling agile locomotion behaviors that adapt to different terrains and disturbances~\cite{peng2018deepmimic,peng2020learning}.
    Subsequent works introduce actuator models and domain randomization, and teacher-student frameworks to enable robust sim-to-real transfer~\cite{hwangbo2019learning}.
    Meanwhile, advances in massively parallel simulation have greatly improved sampling efficiency, reducing training duration to mere minutes and supporting more complex control architectures~\cite{makoviychuk2021isaac,kumar2021rma,rudin2022learning,zuo2024learning,chen2024learning}.
    Additional efforts integrate learning-based controllers with traditional control structures, adaptive compensation, or bio-inspired priors to enhance robustness and generalization~\cite{abdi2019reinforcement,yao2021hierarchical,gangapurwala2022rloc,zhang2022learning,bellegarda2022cpg,sun2024learning}.
    While effective in practice, these approaches primarily intervene at the level of architecture design, reward shaping, curriculum scheduling, or inductive bias injection, and do not explicitly modify the structure of the underlying policy optimization problem or its gradient properties.
    In practical legged robot systems, it is also common to impose explicit constraints on control outputs, such as action scaling, saturation, or conservative torque limits, particularly during early training, to mitigate unstable exploration and satisfy hardware constraints~\cite{buchli2009compliant,calanca2015review,kim2023torque}.
    Despite their widespread use, such mechanisms are typically treated as fixed engineering heuristics rather than explicit components of the learning formulation, and their implications from an optimization perspective remain largely unexplored.

%%%%%%%%%%%%%%%%%%%%%%%%%%%%%%%%%%%%%%%%%%%%%%%%%%%%%%%%%%%%%%%%%%%%%%%%%%%%%%%%
%%%%%%%%%%%%%%%%%%%%%%%%%%%%%%%%%%%%%%%%%%%%%%%%%%%%%%%%%%%%%%%%%%%%%%%%%%%%%%%%

\section{Background}

    Learning-based control for legged robot is commonly formulated as a sequential decision-making problem under uncertainty.
    In this setting, a robot interacts with a complex physical environment characterized by intermittent contacts, nonlinear dynamics, and strict actuation constraints.
    Such problems are naturally modeled as a Markov Decision Process (MDP)
    $\langle \mathcal{S}, \mathcal{A}, P, p_0, r, \gamma \rangle$,
    where the state space $\mathcal{S}$ typically includes proprioceptive feedback, contact information, and task-related observations, and the action space $\mathcal{A}$ corresponds to continuous control commands.
    At each timestep $t$, the controller observes a robot state $s_t \in \mathcal{S}$ and outputs an action $a_t \in \mathcal{A}$, which is directly mapped to joint-level commands.
    In this work, we focus on \emph{torque-based control}, where actions represent desired joint torques and are executed under strict hardware and safety limits.
    The environment transitions according to the underlying robot dynamics and contact interactions, modeled by $P(s_{t+1}\,|\,s_t,a_t)$, with the initial state drawn from $p_0$.
    A scalar reward $r(s_t,a_t)$ encodes locomotion objectives such as stability, progress, and energy efficiency, and the discount factor $\gamma \in [0,1)$ balances short-term control accuracy and long-horizon task performance.
    A parameterized stochastic policy $\pi_\theta(a\,|\,s)$ induces trajectories
    $\tau=(s_0,a_0,\ldots,s_T,a_T)$, and policy learning aims to maximize the expected discounted return
    \begin{equation}
        \max_\theta \; \mathbb{E}_{\tau \sim \pi_\theta,P}
        \Bigg[ \sum_{t=0}^{\infty} \gamma^t r(s_t,a_t) \Bigg].
    \end{equation}

    Policy-gradient methods optimize this objective by directly updating the policy parameters based on sampled interaction data.
    Among them, Proximal Policy Optimization (PPO) has become a standard choice for legged robot control due to its empirical stability and simplicity when applied to high-dimensional continuous actions.
    PPO achieves this stability through a clipped surrogate objective,

    \begin{equation}
       \begin{aligned}
            L^{\text{CLIP}}(\theta)
            = \mathbb{E}_t \Big[
            &\min\!\big(
            r_t(\theta)\hat{A}_t, \\
            &\qquad \text{clip}(r_t(\theta), 1-\epsilon, 1+\epsilon)\hat{A}_t
            \big)
            \Big]
        \end{aligned} 
    \end{equation}

    where $r_t(\theta)=\pi_\theta(a_t|s_t)/\pi_{\theta_{\text{old}}}(a_t|s_t)$ denotes the probability ratio between the updated and previous policies, and $\hat{A}_t$ is an advantage estimate.

    Despite its strong empirical performance, PPO implicitly assumes a fixed action space throughout training.
    In high-dimensional torque-based control, this assumption couples early optimization with the full action space, often leading to inefficient exploration and high-variance gradient estimates before a reliable control regime is established.
    Our objective is to improve optimization efficiency across training stages by explicitly modeling how action space should evolve during learning.

    \section{From PPO to GPO}

    To explicitly model the evolution of the action space during learning while retaining PPO’s stable optimization framework, we introduce \emph{Growing Policy Optimization (GPO)}.
    GPO modifies neither the policy architecture nor the reward design.
    Instead, it introduces a time-dependent transformation that maps the policy’s latent action outputs to executable actions with a progressively expanding range.
    This formulation preserves PPO’s update structure while allowing the effective action space to grow smoothly over training.

    \subsection{Action Space Transformation}
    Consider a stochastic policy parameterized by $\theta$, which produces a latent Gaussian action
    $
        a \sim \mathcal{N}(\mu_\theta(s), \sigma^2).
    $
    In standard implementations, action limits are typically enforced by hard clipping
    $a=\text{clip}(a, -a_{\text{limit}}, a_{\text{limit}})$, which introduces a
    non-differentiable truncation and discards gradient information once the action saturates.
    In contrast, GPO enforces action bounds through a smooth, time-dependent transformation.
    Specifically, GPO applies
    \begin{equation}
        \tilde{a} = \beta_t \tanh\!\left(\tfrac{a}{\beta_t}\right),
    \end{equation}
    where $\beta_t = a_{\text{limit}} f(t)$ is a monotonically increasing growth function of
    the training step $t \in (0,T]$, with $\lim_{t\to\infty} f(t)=1$.
    This transformation induces a bounded yet progressively expanding effective action space.
    
    Notably, when $\beta_t$ approaches its maximum value $a_{\text{limit}}$, the transformation
    recovers the behavior of standard PPO with action clipping.
    Specifically, for small actions satisfying $|a| \ll \beta_t$, the transformation operates in a
    near-linear regime,
    $
        \tilde{a}
        = \beta_t \tanh\!\left(\tfrac{a}{\beta_t}\right)
        \approx a,
    $
    while for large-magnitude actions, the output smoothly saturates at the action limit,
    $
        \tilde{a} \approx \pm a_{\text{limit}},
    $
    which is functionally equivalent to hard clipping but preserves smoothness.

    Throughout the analysis, we assume that the scaled action satisfies
    $\tfrac{a}{\beta_t} \in [-0.5, 0.5]$, ensuring that the transformation predominantly operates in
    the near-linear regime during optimization.
    From an engineering perspective, robots are rarely operated under full-load conditions, and thus
    the scaled action typically satisfies $\tfrac{a}{\beta_t} \in [-0.5, 0.5]$ in practice, which we
    also verify empirically in our experiments.

    \subsection{Effect of action space Growth}
    
    Under GPO, the policy interacts with the environment through the transformed actions $\tilde{a}$.
    As the range parameter $\beta_t$ varies over training, the same policy network output corresponds to different executable actions.
    This introduces a time-varying action transformation into the optimization process, which may influence
    both the policy-gradient update and its stability.
    We first show that the GPO transformation preserves the PPO update rule up to a change of variables.
    \paragraph{Update Equivalence.}
    The importance sampling ratio under GPO satisfies
    \begin{equation}
        \begin{aligned}
            r_t^{\text{GPO}}(\theta)
            &= \frac{\pi_\theta(\tilde{a}_t \mid s_t)}{\pi_{\theta_{\mathrm{old}}}(\tilde{a}_t \mid s_t)} \\
            &= \frac{\pi_\theta(a_t \mid s_t)}{\pi_{\theta_{\mathrm{old}}}(a_t \mid s_t)} = r_t^{\text{PPO}}(\theta).
        \end{aligned}
    \end{equation}
    implying that the clipped surrogate objective optimized by GPO is structurally identical to that of PPO.
    
    While the update rule is preserved, the action transformation modifies the log-likelihood through an additional Jacobian term.
    We therefore further quantify how this transformation affects the policy gradient itself.

    \paragraph{Gradient Difference Bound.}
    Assume $\tfrac{a}{\beta_t}\in[-0.5,0.5]$.
    Then, for the action $\tilde a$, the difference between the gradients
    induced by GPO and PPO satisfies
    \begin{equation}
        \bigl\|
        \nabla_\theta \log \pi_{\theta}(\tilde a)
        -
        \nabla_\theta \log \pi_{\theta}(a)
        \bigr\| \le\ C\|\nabla_\theta \mu\|,
    \end{equation}
    where $C=\tfrac{\sinh(1)-1}{2\sigma^2}|\beta_{\max}-\beta_t|$.
    This bound shows that, despite the time-varying action transformation introduced by GPO, the resulting gradient distortion remains bounded and decreases monotonically as $\beta_t$ increases.
    Consequently, GPO preserves PPO’s update structure while maintaining gradient stability throughout training. A detailed derivation is provided in Appendix~\ref{app:proofs0}.
    
    \subsection{Early-Stage Optimization: Signal-to-Noise Ratio and Local Convergence}

    Having established that GPO preserves PPO’s update structure while introducing only bounded gradient distortion,
    we now analyze how the action space parameter $\beta_t$ influences stochastic optimization behavior in the early stage of training.
    In this regime, learning dynamics are dominated by gradient noise, local convergence speed, and the quality of the resulting control policy.
    
    Let $g_t$ denote a single-sample policy-gradient estimator under GPO,
    \begin{equation}
        g_t = \nabla_\theta \log \pi_\theta(\tilde{a}_t \mid s_t)\, A_t,
    \end{equation}
    where $A_t$ is an advantage estimate.
    We first characterize how the variance of this estimator depends on the effective action space.
    
    \paragraph{Gradient Variance.}
    Under standard regularity assumptions, the variance of the policy-gradient estimator satisfies
    \begin{equation}
        \operatorname{Var}\!\left[g_t\right]
        \leq
        c\,\beta_t^2,
        \quad
        c := \sigma_A^2 K^2,
        \quad
        K = \frac{2}{\sigma^2}\left\|\nabla_\theta \mu_\theta\right\|.
    \end{equation}
    This result shows that gradient variance grows quadratically with the action space parameter $\beta_t$.
    Consequently, restricting the effective action space during early training significantly reduces stochastic gradient noise and stabilizes optimization.
    
    \paragraph{Signal-to-Noise Ratio.}
    Since gradient variance directly affects the reliability of stochastic updates, we examine its impact on the signal-to-noise ratio (SNR).
    The SNR of the policy-gradient estimator can be approximated as
    \begin{equation}
        \operatorname{SNR}(\beta_t)
        \;\approx\;
        \frac{S_0}{\sqrt{c}} \cdot \frac{1}{\beta_t},
        \quad
        S_0 := \bigl\|\mathbb{E}[g_t]\bigr\|_{\beta_t=\beta_{\min}} .
    \end{equation}
    This inverse dependence implies that smaller action spaces yield higher SNR, providing a cleaner learning signal and enabling more reliable parameter updates in early training.
    
    \paragraph{Convergence Error Bound.}
    To formalize how reduced gradient noise affects early-stage learning, consider $J(\theta)$ is the expected accumulated return and the parameter $\theta^\star$ satisfying $\nabla J(\theta^\star)=0$ corresponds to a stationary policy, which represent a local optimum of the return.
    Under a fixed step size $\eta \leq \mu/L^2$, the expected squared distance to the local optimum satisfies
    \begin{equation}
    		\mathbb{E}\left\|\theta_t-\theta^*\right\|^2 \leq(1-\eta \mu)^t\left\|\theta_0-\theta^*\right\|^2+\frac{\eta}{\mu} c \beta_t^2.
    \end{equation}	
    The transient term decays and the steady-state error is dominated by the variance term, which scales with $\beta_t^2$.
    As a result, a smaller action space leads to closer proximity to the local optimum within the same number of training steps.

    \paragraph{Early-Stage Return Advantage.}
    We further show that restricting the effective action space improves local convergence
    and leads to superior early-stage performance.
    Consider two training protocols with the same step size $\eta$:
    (i) GPO, which uses a smaller effective action space $\beta_t$ during early training
    and gradually expands it;
    (ii) a fixed-range baseline that always uses $\beta_{\max}$.
    Let $\{\theta_t\}$ and $\{\bar\theta_t\}$ denote the corresponding parameters.
    
    Under standard smoothness and local strong convexity assumptions,
    after $T_1$ training steps the expected distance to a local optimum $\theta^\star$ satisfies
    \begin{equation}
    \begin{aligned}
    \mathbb{E}\|\theta_{T_1}-\theta^\star\|
    &\le
    \varepsilon+\sqrt{\tfrac{\eta c}{\mu}}\,\beta_{T_1},\\
    \mathbb{E}\|\bar\theta_{T_1}-\theta^\star\|
    &\le
    \varepsilon+\sqrt{\tfrac{\eta c}{\mu}}\,\beta_{\max},
    \end{aligned}
    \end{equation}
    where $\beta_{T_1}<\beta_{\max}$ and $\varepsilon$ captures the decayed transient term,
    and the expected return after $T_1$ steps satisfies
    \begin{equation}
    \begin{aligned}
    \mathbb{E}[J(\theta_{T_1})]
    &\ge
    J(\theta^\star)
    -\frac{\mu}{2}\!\left(\varepsilon^2+\frac{\eta c}{\mu}\beta_{T_1}^2\right),\\
    \mathbb{E}[J(\bar\theta_{T_1})]
    &\ge
    J(\theta^\star)
    -\frac{\mu}{2}\!\left(\varepsilon^2+\frac{\eta c}{\mu}\beta_{\max}^2\right),
    \end{aligned}
    \end{equation}
    Therefore, GPO achieves a strictly tighter lower bound on the expected return
    in the early stage of training.
    A detailed derivation is provided in Appendix~\ref{app:Faster_Local}.
    
    Taken together, these results show that restricting the effective action space during early training not only stabilizes optimization through higher SNR and lower gradient variance, but also enables faster convergence to locally optimal control policies and superior early-stage task performance.

\subsection{Late-Stage Behavior: Exploration and Asymptotic Performance}

We now analyze the long-term behavior of GPO and show that gradually expanding the effective action space leads to no-worse, and potentially improved, asymptotic performance.
We focus on the regime where the policy parameters remain in a neighborhood of a stationary point $\theta^\star$ satisfying $\nabla J(\theta^\star)=0$.
Within this region, the expected return is locally well approximated by a quadratic function,
\begin{equation}
J(\theta)
=
J(\theta^\star)
-\tfrac{1}{2}(\theta-\theta^\star)^\top H(\theta-\theta^\star),
\end{equation}
where $H$ denotes the local curvature matrix, assumed to be positive definite.
This assumption is standard in local convergence analysis and is consistent with the smoothness conditions used throughout the paper.
Under stochastic policy-gradient updates, the parameter evolution can be written as
$
\theta_{t+1}
=
\theta_t+\eta\bigl(\nabla J(\theta_t)+\xi_t\bigr),
$
where $\eta>0$ is the step size and $\xi_t$ denotes the stochastic gradient noise.
We assume the noise is unbiased, $\mathbb E[\xi_t]=0$, and that its conditional second moment scales with the effective action space,
\begin{equation}
\mathbb E[\|\xi_t\|^2]\le c\beta_t^2\quad \text{(implied by }\mathrm{Var}[g_t]\le c\beta_t^2\text{)}.
\end{equation}
Here, $\beta_t$ is the action space parameter introduced by GPO.
For a fixed range baseline, $\beta_t\equiv\beta_{\max}$ and the corresponding noise variance is uniformly bounded by $c\,\beta_{\max}^2$.
When the step size is sufficiently small, the induced error dynamics around $\theta^\star$ form a stable linear stochastic recursion.
As shown in Appendix~\ref{exploration_behavior}, this yields a steady-state bound on the mean-squared parameter error:
\begin{equation}
\limsup_{t\to\infty}\mathbb E\|\theta_t-\theta^\star\|^2
\;\le\;
\frac{\eta^2 c}{1-\rho}\,\beta_\infty^2,
\end{equation}
where $\rho\in(0,1)$ is a contraction factor determined by the local curvature and step size.
For the fixed-range baseline, the same bound holds with $\beta_\infty$ replaced by $\beta_{\max}$.
Since $\beta_\infty\le\beta_{\max}$ by construction, GPO achieves a no-worse steady-state parameter error.
Finally, the local quadratic form of $J(\theta)$ implies that smaller steady-state parameter error directly translates into higher expected return.
Consequently,
\begin{equation}
\liminf_{t\to\infty}\mathbb E[J_{\mathrm{GPO}}]
\;\ge\;
\liminf_{t\to\infty}\mathbb E[J_{\mathrm{fixed}}] .
\end{equation}

The late-stage analysis shows that gradually expanding the effective action space can improve asymptotic expected performance.
Combined with the early-stage analysis, GPO achieves a favorable trade-off:
stable and efficient learning in the early stage, without sacrificing long-term asymptotic performance.
A complete derivation is provided in Appendix~\ref{exploration_behavior}.

\section{Experiments}
\subsection{Tasks and Experimental Setup}
We evaluate the effectiveness of GPO on two torque-controlled locomotion problems:
\emph{quadruped whole-body control} and \emph{hexapod locomotion control}.
To ensure a consistent learning setup across tasks, we adopt the same problem
formulation and training framework, following the asynchronous actor-critic
architecture proposed in \textsc{DreamWaQ}~\cite{nahrendra2023dreamwaq}.
Compared to position control, \emph{torque-level} control removes low-level servo
stabilization and directly exposes contact-rich dynamics to policy learning,
thereby imposing stricter requirements on exploration efficiency and closed-loop
stability.
At each time step, the actor conditions on the observation $o_t$ together with an
estimator output $e_t$, and directly outputs joint torques(see Appendix~\ref{app:implementation} for details).
The observation vector is defined as
$o_t = [w_t,\, g_t,\, q,\, \dot{q},\, v_{\text{cmd}},\, \tau,\, \zeta_t]^{\top}$, 
where $o_t$ combines standard proprioceptive signals, task commands, torque feedback,
and a fatigue-related state following \textsc{SATA}~\cite{li2025sata}.
The command $v_{\text{cmd}}$ differs across tasks, reflecting task-specific tracking
objectives. And we use Gompertz growth curve $\beta_t=e^{-e^{-k\left(t-t_0\right)}} \label{eq:growth}$ as GPO growth function.
\begin{figure}[htpb]
      \centering
        \subfloat{
            \includegraphics[width=0.23\textwidth]{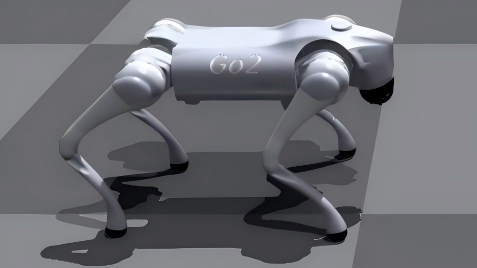}
        }
        \hspace{-0.2cm}
        \subfloat{
            \includegraphics[width=0.23\textwidth]{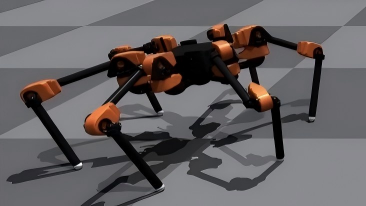}
        }
        \caption{Simulated quadruped (left) and hexapod robot (right).}
        \label{Tasks}
\end{figure}
\paragraph{Reward Terms.}
Following \textsc{SATA}~\cite{li2025sata} and \textsc{HeLom}~\cite{yang2025helom}, we adopt a unified reward structure across
both tasks, consisting of shared task-agnostic \emph{regularization} terms and
\emph{task-specific tracking} terms.
The regularization encourages torque efficiency, smooth control, stable body motion,
and safe contact interactions, while task reward differs in the tracking objectives: for the quadruped, it includes planar velocities $(v_x, v_y)$, yaw rate $\omega_z$, target base height, and target pitch angle; for the hexapod, it includes $(v_x, v_y, \omega_z)$ only.
We \emph{don't} impose any explicit gait priors, such as phase synchronization,
predefined footfall schedules, or leg-usage constraints.
Instead, locomotion behaviors \emph{emerge} solely from optimizing task tracking under
shared task-agnostic regularization in torque-level, contact-rich dynamics.
The full reward is provided in Appendix~\ref{app:reward} Tab.~\ref{tab:rewards}.

    \paragraph{Task Focus and Evaluation Criteria}
    For the quadruped whole-body control task, we emphasize closed-loop stability and robustness while simultaneously tracking commanded motion and regulating body posture (e.g., target height and pitch), and we assess disturbance rejection under external pushes. 
    In contrast, for the hexapod locomotion task we focus on the emergence of \emph{coherent and sustainable} coordination patterns. Due to higher actuation redundancy and a larger feasibility set for maintaining balance, a hexapod policy may converge to a simplified local optimum when no explicit gait-structure or load-balancing incentives are imposed, exhibiting \emph{degenerate support allocation} where propulsion and support are overly concentrated on a subset of legs while the remaining legs are under-utilized. Accordingly, we evaluate the rationality of support distribution and the regularity and transferability of the resulting coordination patterns, alongside standard tracking and efficiency metrics.
    \subsection{Effect of Action Space Growth Functions}
        We compare several representative action space growth functions
        (Tab.~\ref{tab:growth_schedules}), including no growth (PPO), linear with clipping,
        sigmoid, and the proposed Gompertz (GPO).
        These functions induce distinct growth-rate behaviors: constant (linear),
        monotonically decreasing (sigmoid), and asymmetric with slow initialization and
        accelerated expansion (Gompertz).
        
\begin{table}[htbp]
  % \vspace{-0.4cm}
  \caption{Action space growth functions.}
  \label{tab:growth_schedules}
  % \vspace{-0.05in}
  \begin{center}
    \scriptsize
    \setlength{\tabcolsep}{2.5pt}
    \renewcommand{\arraystretch}{0.9}
    \begin{sc}
      \begin{tabular}{lcc}
        \toprule
          & \textbf{Growth Function $\beta_t$} & \textbf{Parameters} \\
        \midrule
        No growth
        & $1$
        & -- \\

        Linear
        & $\mathrm{clip}(k t, 0, 1)$
        & $k=\frac{1}{3000}$ \\

        Sigmoid
        & $(1+\exp(-k(t-t_0)))^{-1}$
        & $k=-2.3{\times}10^{-3},\ t_0=3{\times}10^{3}$ \\

        Gompertz
        & $e^{-e^{-k(t-t_0)}}$
        & $k=3{\times}10^{-5},\ t_0=2.4{\times}10^{4}$ \\
        \bottomrule
      \end{tabular}
    \end{sc}
  \end{center}
  \vskip -0.1in
  % \vspace{-0.4cm}
\end{table}
    
        \begin{figure}[htbp]
            \includegraphics[width=0.46\textwidth]{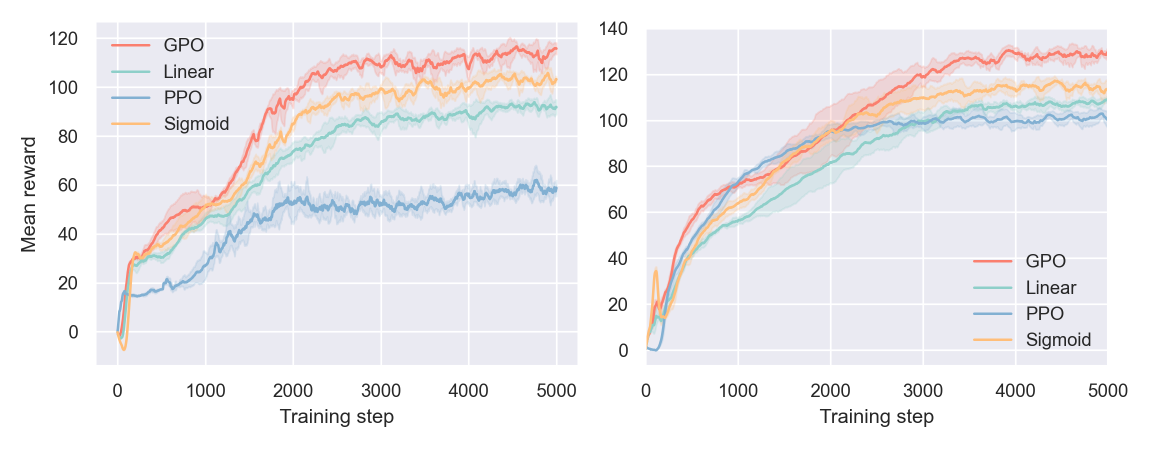}
            % \setlength{\abovecaptionskip}{2pt}
            % \vspace{-0.2cm}
            \caption{Training rewards under different growth functions.
            Left: quadruped whole-body.
            Right: hexapod locomotion.}
            \label{growth_curve}
        \end{figure}
        \begin{figure*}[t]
            \centering
            \includegraphics[width=\textwidth]{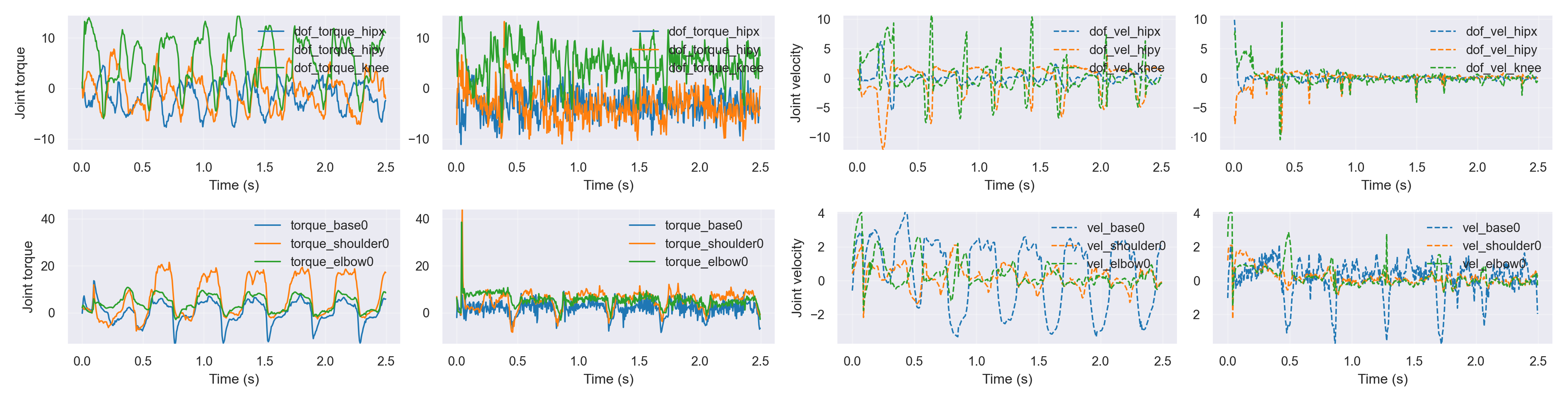}
            % \vspace{-0.5cm}
            \caption{
                \textbf{Joint-level simulation results for the front-left leg.}
                Top row: quadruped. Bottom row: hexapod.
                For each robot, columns from left to right show
                (1) joint torque with GPO,
                (2) joint torque with PPO,
                (3) joint velocity with GPO,
                and (4) joint velocity with PPO.
            }
            \label{fig:single_leg_joint_dynamics}
        \end{figure*}
        As shown in Fig.~\ref{growth_curve}, the Gompertz function consistently achieves the best performance across both tasks.
        In the early stage of training, GPO exhibits faster reward improvement and more stable learning behavior,
        while PPO with a fixed action space converges significantly more slowly, especially on the more challenging task.
        This indicates that allowing the policy to explore the full action space from the beginning leads to an overly large exploration space,
        making it difficult for the policy to acquire effective control behaviors.
        Among the growth functions, sigmoid growth improves stability compared to linear growth,
        but converges more slowly and attains a lower final reward.
        Linear growth performs the worst overall, often exhibiting unstable training dynamics or premature performance saturation,
        suggesting that a constant growth rate is insufficient to match the learning requirements.
        
        These results highlight the importance of asymmetric action space growth.
        In the early stage, a smaller initial growth slope imposes a stronger restriction on the action space,
        which encourages the policy to learn reliable low-amplitude control behaviors within a limited exploration space.
        In later stages, increasing the growth rate progressively restores control authority and enlarges the exploration space, allowing the policy to refine behaviors and escape suboptimal situation.
        This explains the advantage of the Gompertz function, which effectively balances early-stage learnability with late-stage performance.
    \subsection{Simulation Experiment}
        We further analyze the performance of our methods in simulation under a set of fixed command targets. A total of 500 command samples are randomly generated for evaluation. For the quadruped robot, we evaluate tracking performance with respect to both velocity and body pose commands, including $(v_x, v_y, \omega_z, H_b, \theta_{\text{pitch}})$, while for the hexapod robot, the evaluation focuses on $(v_x, v_y, \omega_z, H_b)$.

\begin{table}[htbp]
  % \vspace{-0.4cm}
  \caption{Tracking error statistics (mean $\pm$ half-range).}
  \label{tab:gpo_ppo_tracking}
  % \vspace{-0.05in}
  \begin{center}
    \scriptsize
\setlength{\tabcolsep}{1pt}
\renewcommand{\arraystretch}{0.9}
\begin{sc}
\begin{tabular}{lcccc}
\toprule
& \multicolumn{2}{c}{Whole-body control}
& \multicolumn{2}{c}{Velocity tracking} \\
\cmidrule(lr){2-3}\cmidrule(lr){4-5}
& GPO & PPO & GPO & PPO \\
\midrule
\textbf{$v_x, v_y${\normalfont (m/s)}}  & $0.015 \pm 0.035$ & $-0.30 \pm 0.10$
                         & $0.00 \pm 0.05$   & $-0.10 \pm 0.05$ \\
\textbf{$\omega_z${\normalfont (rad/s)}} & $0.00 \pm 0.05$ & $0.00 \pm 0.05$
                           & $0.00 \pm 0.05$ & $0.02 \pm 0.05$ \\
\textbf{$H_b${\normalfont (m)}} & $0.005 \pm 0.0025$ & $-0.03 \pm 0.02$
                          & $0.02 \pm 0.005$  & $-0.12 \pm 0.03$ \\
\textbf{$\theta_{\text{\normalfont{pitch}}}${\normalfont (rad)}}     & $0.01 \pm 0.06$ & $0.00 \pm 0.20$ & -- & -- \\
\bottomrule
\end{tabular}
\end{sc}
  \end{center}
  % \vskip -0.1in
  % \vspace{-0.4cm}
\end{table}

\begin{figure}[tbp]
            \includegraphics[width=0.45\textwidth]{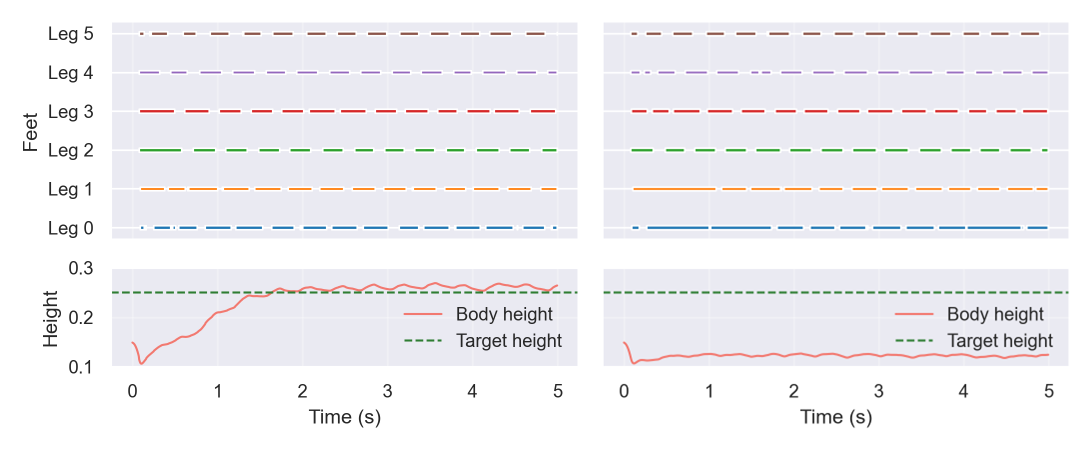}
            % \setlength{\abovecaptionskip}{2pt}
            % \vspace{-0.2cm}
            \caption{\textbf{Hexapod gait and body height visualization.}
            Left: GPO; right: PPO. For each method, the top shows six-leg contact patterns,
            and the bottom compares target and actual body height.}
            \label{gait}
    \end{figure}

    As shown in Tab.~\ref{tab:gpo_ppo_tracking}, GPO consistently exhibits more accurate tracking performance than PPO. For the quadruped robot, PPO fails to achieve effective velocity tracking even along the $x$ direction alone.
    A similar behavior is observed on the hexapod robot. Under the same desired velocity commands, PPO exhibits tracking errors on the order of $0.1~\mathrm{m/s}$, while GPO consistently achieves closer adherence to the target velocities. Furthermore, as shown in Fig.~\ref{gait}, PPO demonstrates degraded performance in tracking fixed body-height commands. When the desired body height is set to $0.25~\mathrm{m}$, the resulting steady-state height achieved by PPO remains below $0.15~\mathrm{m}$, indicating a substantial deviation from the target. Such limitations may be associated with the lack of progressive increases in torque limits or effective policy execution dynamics, which are implicitly addressed by GPO.
    
    Joint-level analysis reveals a clear qualitative difference between the two optimization schemes. As shown in Fig.~\ref{fig:single_leg_joint_dynamics}, GPO-trained policies exhibit pronounced periodicity in joint velocities and torques during locomotion, indicating the emergence of coordinated and repeatable motion patterns. PPO-trained policies, in contrast, produce less regular and more erratic joint behaviors, and are more prone to uneven leg utilization when no explicit gait-structure or load-balancing incentives are imposed.
    Here, the action limits are set to $a_{\text{limit}}=32$ for the quadruped and $a_{\text{limit}}=40$ for the hexapod; as illustrated by the joint torque plots in Fig.~\ref{fig:single_leg_joint_dynamics}, the resulting actions remain well within the range $\tfrac{a}{\beta_t}\in[-0.5,,0.5]$, thereby satisfying the assumption used in our theoretical analysis.
    This observation is further supported by the hexapod gait diagram in Fig.~\ref{gait}. Compared with the variant without action space growth, the policy trained with growth yields a more periodic and stable stepping rhythm, with more consistent stance swing alternation and inter-leg coordination, and fewer signs of degenerate support allocation, such as prolonged ground support dominated by the forelegs. These results suggest that action space growth facilitates the formation of structured coordination patterns rather than task-specific behaviors.
    
	\subsection{Hardware Validation}
    To validate the effectiveness of GPO, we deploy the learned policy on a Unitree Go2 quadruped robot and compare it against several representative baselines, including torque-based PPO, Pure Torque DeCAP~\cite{sood2024decap}, and Position-assisted DeCAP. We first evaluate GPO on training-seen whole-body control tasks to verify stable locomotion and posture regulation under nominal conditions. In addition, we apply human-induced perturbations by manually lifting robot legs. Under these disturbances, GPO enables the robot to respond naturally to human-applied forces, adjusting its body posture smoothly without triggering unnecessary stepping or recovery motions.
    We further assess robustness in multiple unseen scenarios that are not encountered during training. 
    \begin{figure}[tbp]
        \includegraphics[width=\linewidth]{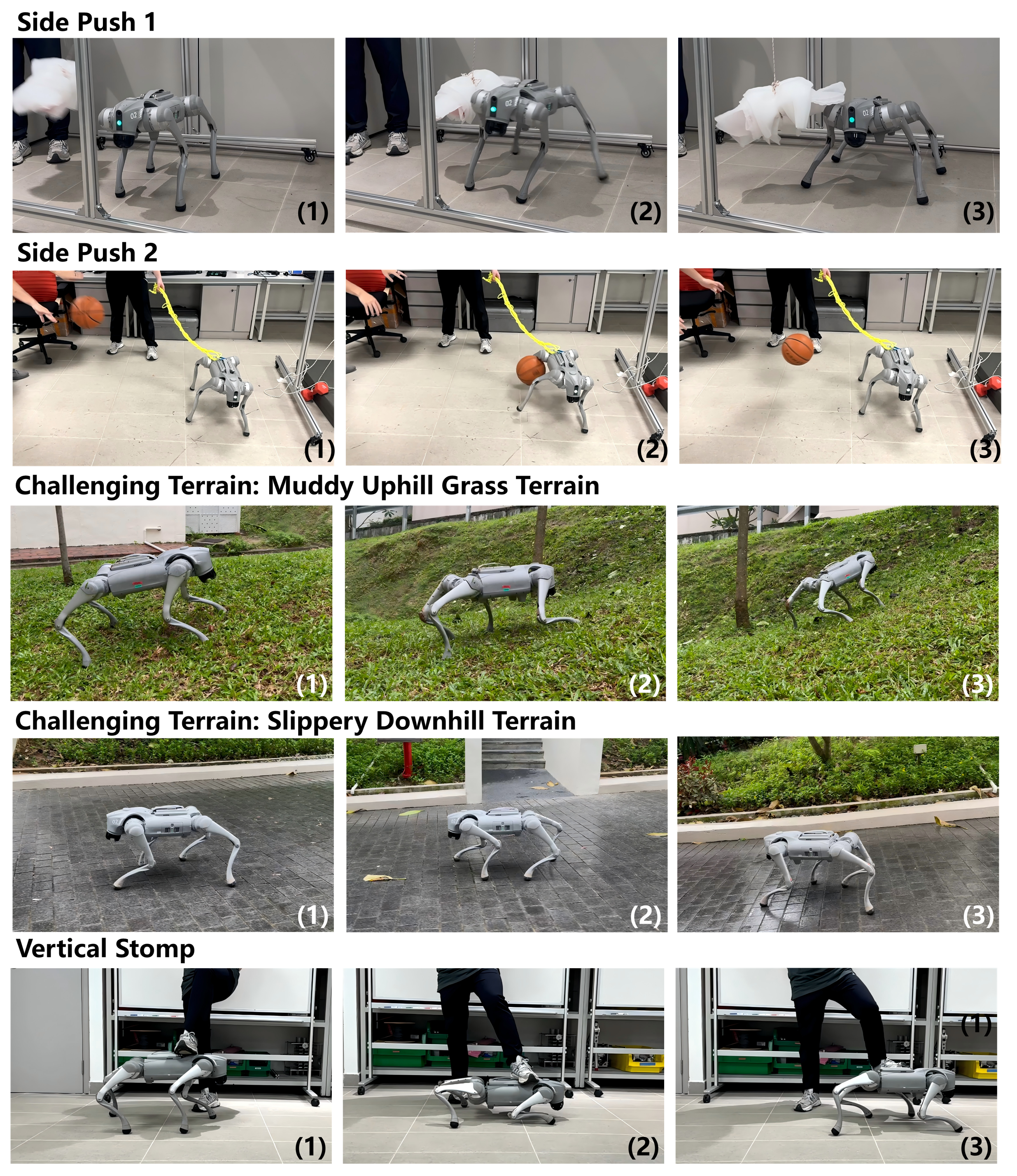}
        \setlength{\abovecaptionskip}{2pt}
        \vspace{-0.6cm}
        \caption{Hardware validation under disturbance tasks. GPO demonstrates robust locomotion and strong recovery capability under external disturbances.}
        \label{fig:hardware_task}
    \end{figure}
    \begin{table}[tbp]
    \vspace{-0.3cm}
      \caption{Robustness performance comparison across different perturbation tests (success rate over 10 trials).}
      \label{tab:performance_comparison}
      \vspace{-0.05in}
      \begin{center}
        \footnotesize
        \setlength{\tabcolsep}{3pt}
        \renewcommand{\arraystretch}{0.95}
        \begin{sc}
          \begin{tabular}{lccc}
              \toprule
              & \begin{tabular}[c]{@{}c@{}}Side\\Push\end{tabular}
              & \begin{tabular}[c]{@{}c@{}}Challenging\\Terrain\end{tabular}
              & \begin{tabular}[c]{@{}c@{}}Vertical\\Stomp\end{tabular} \\
              \midrule
              \textbf{GPO(Ours)} & \textbf{100\%} & \textbf{100\%} & \textbf{100\%} \\
              \textbf{PPO} & 0\% & 20\% & 40\% \\
              \textbf{DeCAP(Torque)} & 0\% & 0\% & 60\% \\
              \textbf{DeCAP(Position)} & 80\% & 40\% & 70\% \\
              \bottomrule
          \end{tabular}
        \end{sc}
      \end{center}
      \vskip -0.1in
      \vspace{-0.3cm}
    \end{table}
    These tests include \textbf{side push} disturbances induced by a dropped 5 kg dumbbell and thrown basketballs, \textbf{challenging terrain} locomotion over muddy uphill grass terrain and slippery downhill terrain, and \textbf{vertical stomp}, where the robot walks under external downward forces or presses. As shown in Fig.~\ref{fig:hardware_task}, 
    in the side push and vertical stomp tasks, GPO exhibits strong recovery capabilities against sudden external disturbances, allowing the robot to regain balance without relying on any predefined instructions before operation.
    For the challenging terrain task, after experiencing slips during locomotion, the robot is able to rapidly recover a stable standing posture. The GPO controller dynamically adjusts motor outputs to stabilize the robot, without invoking predefined recovery motions.
    Evaluation results are summarized in Table~\ref{tab:performance_comparison}. Across all tests, GPO consistently outperforms the baseline methods(please refer to our associated video for more details).
    
	\section{Conclusion}
    In this work, we introduced GPO, a training framework that explicitly models the evolution of the effective action space during reinforcement learning. By applying a time-varying action transformation, GPO preserves standard PPO updates while improving gradient stability across training stages. This formulation leads to faster convergence in the early phase and higher expected return in the later phase, resulting in improved overall training efficiency.
    We evaluated GPO on torque-controlled locomotion and whole-body control tasks across both quadruped and hexapod robots. 
    Results from simulations and real-world hardware experiments demonstrate that GPO consistently improves performance and robustness, including under external disturbances and zero-shot sim-to-real transfer.
    Overall, GPO provides a general, environment-agnostic optimization framework for high-dimensional continuous control, offering a practical training paradigm for legged robots operating in contact-rich and uncertain real-world environments.

%%%%%%%%%%%%%%%%%%%%%%%%%%%%%%%%%%%%%%%%%%%%%%%%%%%%%%%%%%%%%%%%%%%%%%%%%%%%%%%%
%%%%%%%%%%%%%%%%%%%%%%%%%%%%%%%%%%%%%%%%%%%%%%%%%%%%%%%%%%%%%%%%%%%%%%%%%%%%%%%%

% \newpage
\bibliographystyle{IEEEtran}
\bibliography{IEEEexample}

%%%%%%%%%%%%%%%%%%%%%%%%%%%%%%%%%%%%%%%%%%%%%%%%%%%%%%%%%%%%%%%%%%%%%%%%%%%%%%%%
%%%%%%%%%%%%%%%%%%%%%%%%%%%%%%%%%%%%%%%%%%%%%%%%%%%%%%%%%%%%%%%%%%%%%%%%%%%%%%%%

\onecolumn
\appendix
\section{Effect of action space Growth}
	\label{app:proofs0}

    \subsection{Update Equivalence}
    \label{app:update_equivalence}
    
        We begin by recalling the clipped surrogate objective used in PPO,
        \begin{equation}
        L^{\text{CLIP}}(\theta)
        =
        \mathbb{E}_t \!\left[
        \min\!\left(
        r_t(\theta)\,\hat{A}_t,\;
        \text{clip}(r_t(\theta), 1-\epsilon, 1+\epsilon)\,\hat{A}_t
        \right)
        \right],
        \label{eq:A3_ppo}
        \end{equation}
        where the importance sampling ratio is defined as
        \begin{equation}
        r_t(\theta)
        =
        \frac{\pi_\theta(a_t \mid s_t)}{\pi_{\theta_{\mathrm{old}}}(a_t \mid s_t)}.
        \label{eq:A3_ratio}
        \end{equation}
        
        Under GPO, the executed action is obtained via the transformation
        \(
        \tilde{a} = \beta_t \tanh(a / \beta_t)
        \).
        Substituting $\tilde{a}$ for $a$ yields the modified ratio
        \begin{equation}
        r_t^{\text{GPO}}(\theta)
        =
        \frac{\pi_\theta(\tilde{a}_t \mid s_t)}{\pi_{\theta_{\mathrm{old}}}(\tilde{a}_t \mid s_t)}.
        \label{eq:A3_r1}
        \end{equation}
        
        By the change-of-variables formula, the transformed density satisfies
        \begin{equation}
        \pi_\theta(\tilde{a} \mid s_t)
        =
        \pi_\theta(a \mid s_t)
        \left|
        \frac{d a}{d \tilde{a}}
        \right|.
        \label{eq:A3_pi_change}
        \end{equation}
        Since the inverse transformation is given by
        \begin{equation}
        a
        =
        \beta_t \tanh^{-1}\!\left( \tfrac{\tilde{a}}{\beta_t} \right),
        \label{eq:A3_inverse}
        \end{equation}
        the Jacobian determinant evaluates to
        \begin{equation}
        \frac{d \tilde{a}}{d a}
        =
        \text{sech}^2\!\left( \tfrac{a}{\beta_t} \right)
        =
        1 - \tanh^2\!\left( \tfrac{a}{\beta_t} \right)
        =
        1 - \left( \tfrac{\tilde{a}}{\beta_t} \right)^2.
        \label{eq:A3_jacobian}
        \end{equation}

        For a $d$-dimensional action space, this yields
        \begin{equation}
        \pi_\theta(\tilde{a} \mid s_t)
        =
        \pi_\theta(a \mid s_t)
        \prod_{i=1}^d
        \left(
        1 - \left( \tfrac{\tilde{a}_i}{\beta_t} \right)^2
        \right)^{-1}.
        \label{eq:A3_density}
        \end{equation}
        Define the Jacobian term
    	\begin{equation}
    		J(\theta) = \prod_{i=1}^d \left( 1 - \left( \frac{\tilde{a}_i}{\beta_t} \right)^2 \right)^{-1}
    	\end{equation}
    	During PPO updates, $\tilde{a}_i$ is fixed for the sampled transition, so the Jacobian term cancels:
    	\begin{align}
    		r_{t}^{GPO}(\theta) = \frac{\pi_\theta(\tilde{a}_t | s_t)}{\pi_{\theta_{\text{old}}}(\tilde{a}_t | s_t)} = \frac{\pi_\theta(a_t | s_t) J(\theta)}{\pi_{\theta_{\text{old}}}(a_t | s_t)J(\theta_{old})} =  \frac{\pi_\theta(a_t | s_t)}{\pi_{\theta_{\text{old}}}(a_t | s_t)}=r_t^{PPO}(\theta)
    		\label{eq:A3_ratio_equal}
    	\end{align}
    
        Equation~\eqref{eq:A3_ratio_equal} shows that the likelihood ratio—and therefore the clipped PPO objective—remains unchanged under the GPO transformation.

%%%%%%%%%%%%%%%%%%%%%%%%%%%%%%%%%%%%%%%%%%%%%%
\subsection{Gradient Difference Bound}
\label{app:gradient_bound}

PPO uses the likelihood-ratio gradient
$\nabla_\theta \log \pi_\theta(a\mid s)$.
In GPO, the environment action is
$\tilde a=h_\beta(a)=\beta\tanh(a/\beta)$,
and the PPO gradient can be written as
$\nabla_\theta \log \pi_{\theta,\beta_{\max}}(\tilde a\mid s)$.
We therefore compare score-function gradients induced by different range
parameters $\beta_t$ and $\beta_{\max}$.

Let $a\sim\mathcal N(\mu_\theta(s),\sigma^2)$ with $\sigma$ independent of $\theta$.
The induced density of $\tilde a$ is
\begin{equation}
\pi_{\theta,\beta}(\tilde a\mid s)
=
\mathcal N\!\big(h_\beta^{-1}(\tilde a);\mu_\theta(s),\sigma^2\big)
\left|\frac{d}{d\tilde a}h_\beta^{-1}(\tilde a)\right|,
\end{equation}
where $h_\beta^{-1}(\tilde a)=\beta\,\operatorname{arctanh}(\tilde a/\beta)$ and
$\left|\frac{d}{d\tilde a}h_\beta^{-1}(\tilde a)\right|=(1-\tilde a^2/\beta^2)^{-1}$
is $\theta$-independent. Hence,
\begin{equation}
\nabla_\theta \log \pi_{\theta,\beta}(\tilde a\mid s)
=
\frac{h_\beta^{-1}(\tilde a)-\mu_\theta(s)}{\sigma^2}\,
\nabla_\theta \mu_\theta(s).
\label{eq:score_beta}
\end{equation}

\paragraph{Gradient difference.}
Comparing GPO ($\beta_t$) and PPO ($\beta_{\max}$) yields
\begin{align}
\Big\|
\nabla_\theta \log \pi_{\theta,\beta_t}(\tilde a\mid s)
-
\nabla_\theta \log \pi_{\theta,\beta_{\max}}(\tilde a\mid s)
\Big\|
&=
\frac{1}{\sigma^2}
\big|h_{\beta_t}^{-1}(\tilde a)-h_{\beta_{\max}}^{-1}(\tilde a)\big|
\;\big\|\nabla_\theta \mu_\theta(s)\big\|.
\label{eq:grad_diff_reduce}
\end{align}

For $\frac{a}{\beta_t}\in[-0.5,0.5]$, which implies
$|\tilde a|\le \beta_t\tanh(0.5)$.
Define $\phi(\beta)=h_\beta^{-1}(\tilde a)=\beta\,\operatorname{arctanh}(\tilde a/\beta)$.
By the mean value theorem,
\begin{equation}
\big|h_{\beta_t}^{-1}(\tilde a)-h_{\beta_{\max}}^{-1}(\tilde a)\big|
\le
\sup_{\beta\in[\beta_t,\beta_{\max}]}\big|\phi'(\beta)\big|\,
|\beta_{\max}-\beta_t|.
\end{equation}
A direct calculation gives
\begin{equation}
\phi'(\beta)
=
\operatorname{arctanh}(u)-\frac{u}{1-u^2},
\qquad
u:=\tilde a/\beta.
\end{equation}
Under $|u|\le\tanh(0.5)$, the maximum is attained at the boundary, yielding
\begin{equation}
\sup_{|u|\le\tanh(0.5)}
\left|
\operatorname{arctanh}(u)-\frac{u}{1-u^2}
\right|
=
\left|
0.5-\frac{\tanh(0.5)}{1-\tanh^2(0.5)}
\right|
=
\frac{\sinh(1)-1}{2}.
\end{equation}
Substituting into \eqref{eq:grad_diff_reduce}, we obtain the explicit bound
\begin{equation}
\bigl\|
        \nabla_\theta \log \pi_{\theta}(\tilde a)
        -
        \nabla_\theta \log \pi_{\theta}(a)
        \bigr\|
=
\Big\|
\nabla_\theta \log \pi_{\theta,\beta_t}(\tilde a\mid s)
-
\nabla_\theta \log \pi_{\theta,\beta_{\max}}(\tilde a\mid s)
\Big\|
\le
\frac{\sinh(1)-1}{2\sigma^2}\,
|\beta_{\max}-\beta_t|\,
\big\|\nabla_\theta \mu_\theta(s)\big\|.
\label{eq:grad_diff_final}
\end{equation}

Equation~\eqref{eq:grad_diff_final} confirms that the gradient distortion introduced by the GPO transformation is bounded and decreases as the range parameter $\beta_t$ increases, ensuring optimization stability comparable to PPO.

	\section{Early-Stage Optimization: Signal-to-Noise Ratio and Local Convergence}
    \label{app:Faster_Local}
	\subsection{Gradient Variance}
    \label{app:Gradient_Variance}
	The latent action produced by the policy network is $a \sim \mathcal{N}(\mu_\theta(s),\sigma^2)$, one-step policy-gradient $g_t=\nabla_\theta \log \pi_\theta\left(\tilde{a}_t \mid s_t\right) A_t$, where $A_t$ is the advantage with $\mathbb{E}\left[A_t\right]=0$ and $\mathbb{E}\left[A_t^2\right]=\sigma_A^2<\infty$. Assume $\mu_\theta$ is differentiable and $\left\|\nabla_\theta \mu_\theta(s)\right\|<\infty$. We restrict $\beta_t, \frac{a}{\beta_t} \in [-0.5, 0.5]$ :
	\begin{equation}
		\operatorname{Var}\left[g_t\right] \leq c \beta_t^2, \quad 
		c := \sigma_A^2 K^2, \ \ 
		K=\frac{2}{\sigma^2}\left\|\nabla_\theta \mu_\theta(s)\right\|
	\end{equation}

	Proof
	
	\begin{equation}
		\tilde{a}=h(a)=\beta_t \tanh \left(\frac{a}{\beta_t}\right)
	\end{equation}
	
	\begin{equation}
		h^{-1}(\tilde{a})=a=\beta_t \operatorname{arctanh}\left(\frac{\tilde{a}}{\beta_t}\right)
	\end{equation}
	
	Let $q(a)$ be the Gaussian density of $a$ and $p(\tilde{a})$ the density of $\tilde{a}$. By the change of variables formula:
	\begin{equation}
		p(\tilde{a})=q\left(h^{-1}(\tilde{a})\right)\left|\frac{d}{d \tilde{a}} h^{-1}(\tilde{a})\right|=\frac{1}{\sqrt{2 \pi} \sigma} \exp \left(-\frac{\left(\beta_t \operatorname{arctanh}\left(\tilde{a} / \beta_t\right)-\mu_\theta(s)\right)^2}{2 \sigma^2}\right)\left(1-\frac{\tilde{a}^2}{\beta_t^2}\right)^{-1}
	\end{equation}
	
	Hence
	
	\begin{equation}
		 \log \pi_\theta(\tilde{a}\mid s) = \log p(\tilde{a}) =  -\frac{1}{2}\log(2\pi\sigma^2)\;-\;\frac{\big(h^{-1}(\tilde{a})-\mu_\theta(s)\big)^2}{2\sigma^2}\;+\;\log\Big(1-\frac{\tilde{a}^2}{\beta_t^2}\Big)^{-1}
	\end{equation}	
	Only $\mu_\theta(s)$ depends on $\theta$, the Jacobian term is $\theta$-independent.
	
	Differentiate w.r.t. $\theta$ :
	\begin{equation}
		\nabla_\theta \log \pi_\theta(\tilde{a} \mid s)=-\frac{1}{\sigma^2}\left(h^{-1}(\tilde{a})-\mu_\theta(s)\right) \nabla_\theta \mu_\theta(s)+0
	\end{equation}	
	Thus 
 	\begin{equation}
 		\left\|\nabla_\theta \log \pi_\theta(\tilde{a} \mid s)\right\|=\frac{\left|h^{-1}(a)-\mu_\theta(s)\right|}{\sigma^2}\left\|\nabla_\theta \mu_\theta(s)\right\|
 	\end{equation}

	For$|x|<1,|\operatorname{arctanh}(x)|=\frac{1}{2}\left|\ln \frac{1+x}{1-x}\right| \leq \frac{|x|}{1-|x|}$. Fix any $r \in(0,1)$ and let $|\tilde{a}| \leq r \beta_t$, $\left|\mu_\theta(s)\right| \leq M \beta_t$ for some $M \geq 0$. Then
	
	\begin{equation}
		\left|h^{-1}(\tilde{a})\right|=\beta_t\left|\operatorname{arctanh}\left(\tilde{a} / \beta_t\right)\right| \leq \beta_t \frac{|\tilde{a}| / \beta_t}{1-|\tilde{a}| / \beta_t} \leq \beta_t \frac{r}{1-r}
	\end{equation}
	 Hence
	 
	 \begin{equation}
	 	\left|h^{-1}(\tilde{a})-\mu_\theta(s)\right| \leq \beta_t\left(\frac{r}{1-r}+M\right)
	 \end{equation}

	 It follows that
	 
	 \begin{equation}
	 \left\|\nabla_\theta \log \pi_\theta(\tilde{a} \mid s)\right\| \leq \frac{\beta_t}{\sigma^2}\left(\frac{r}{1-r}+M\right)\left\|\nabla_\theta \mu_\theta(s)\right\|= K \beta_t
	\end{equation}
	 
	 where
	 
	 \begin{equation}
	 	K=\frac{1}{\sigma^2}\left(\frac{r}{1-r}+M\right)\left\|\nabla_\theta \mu_\theta(s)\right\|
	 \end{equation}

	 A common safe choice is $r=\frac{1}{2}$ and $M \leq 1$, which recovers the simple constant $K=\frac{2}{\sigma^2}\left\|\nabla_\theta \mu_\theta(s)\right\|$

	Since $g_t=\nabla_\theta \log \pi_\theta\left(\tilde{a}_t \mid s_t\right) A_t$,
	
	\begin{equation}
		\left\|g_t\right\| \leq\left|A_t\right| K \beta_t \quad \Longrightarrow \quad \mathbb{E}\left\|g_t\right\|^2 \leq \mathbb{E}\left[A_t^2\right] K^2 \beta_t^2=\sigma_A^2 K^2 \beta_t^2 
	\end{equation}

	Therefore,
	
	\begin{equation}
		\operatorname{Var}\left[g_t\right]=\mathbb{E}\left\|g_t\right\|^2-\left\|\mathbb{E}\left[g_t\right]\right\|^2 \leq \mathbb{E}\left\|g_t\right\|^2 \leq\left(\sigma_A^2 K^2\right) \beta_t^2= c \beta_t^2 
	\end{equation}

    \subsection{Convergence Error Bound}
\label{app:Convergence_Error_Bound}

$J(\theta)$ is the expected accumulated return and the parameter $\theta^\star$ satisfying
$\nabla J(\theta^\star)=0$ corresponds to a stationary policy, which represents a local optimum
of the return.

\begin{equation}
\mathbb{E}\left\|\theta_t-\theta^*\right\|^2
\le
(1-\eta \mu)^t\left\|\theta_0-\theta^*\right\|^2
+\frac{\eta}{\mu}c \beta_t^2
\end{equation}

Proof.
Let $J$ be $\mu$-strongly convex and $L$-smooth. Consider constant-stepsize SGD:
\begin{equation}
\theta_{t+1}=\theta_t-\eta g_t, \quad
\mathbb{E}\left[g_t \mid \theta_t\right]=\nabla J\left(\theta_t\right), \quad
\mathbb{E}\left[\left\|g_t-\nabla J\left(\theta_t\right)\right\|^2 \mid \theta_t\right] \leq \sigma_t^2 .
\end{equation}

Let $ e_t := \theta_t - \theta^\ast $. Expanding one SGD step gives
\begin{equation}
\left\|e_{t+1}\right\|^2
=
\left\|e_t-\eta g_t\right\|^2
=
\left\|e_t\right\|^2-2 \eta e_t^{\top} g_t+\eta^2\left\|g_t\right\|^2 .
\end{equation}

Take conditional expectation given $\theta_t$ and use unbiasedness:
\begin{equation}
\mathbb{E}\left[\left\|e_{t+1}\right\|^2\mid \theta_t\right]
=
\left\|e_t\right\|^2
-2 \eta e_t^{\top} \nabla J\left(\theta_t\right)
+\eta^2\left(\left\|\nabla J\left(\theta_t\right)\right\|^2
+\mathbb{E}\left[\left\|g_t-\nabla J\left(\theta_t\right)\right\|^2 \mid \theta_t\right]\right).
\end{equation}

Strong convexity yields $e_t^{\top} \nabla J\left(\theta_t\right) \geq \mu\left\|e_t\right\|^2$.
$L$-smoothness yields $\left\|\nabla J\left(\theta_t\right)\right\| \leq L\left\|e_t\right\|$.
Hence
\begin{equation}
\mathbb{E}\left\|e_{t+1}\right\|^2
\le
\underbrace{\left(1-2 \eta \mu+\eta^2 L^2\right)}_{:=\rho}
\mathbb{E}\left\|e_{t}\right\|^2+\eta^2 \sigma_t^2 .
\end{equation}

Take total expectation and unroll the recursion:
\begin{equation}
\mathbb{E}\left\|e_t\right\|^2
\le
\rho^t\left\|e_0\right\|^2
+\eta^2 c \beta_t^2 \sum_{i=0}^{t-1} \rho^i
=
\rho^t\left\|e_0\right\|^2+\frac{\eta^2 c \beta_t^2}{1-\rho} .
\end{equation}

Using $1-\rho=2 \eta \mu-\eta^2 L^2 \geq \eta \mu$ when $\eta \leq \mu / L^2$, we get
\begin{equation}
\mathbb{E}\left\|\theta_t - \theta^\ast\right\|^2
\le
(1-\eta \mu)^t\left\|\theta_0-\theta^*\right\|^2
+\frac{\eta}{\mu}c \beta_t^2.
\end{equation}

\subsection{Early-Stage Return Advantage}
\label{Early_Return}

We compare two training protocols under the same step size $\eta$ but different action spaces:
(i) GPO uses a small range $\beta_{t}$ and gradually expands to $\beta_{\max}$;
(ii) the fixed baseline always uses $\beta_{\max}$.
Denote their parameters by $\{\theta_t\}$ and $\{\bar\theta_t\}$, respectively.

\textbf{In early stage ($0$ to $T_1$ steps, small $\beta_t$):}

By Appendix~\ref{app:Convergence_Error_Bound}, for $0<\eta \leq \mu / L^2$,
\begin{equation}
\mathbb{E}\left\|\theta_{T_1}-\theta^{\star}\right\|^2
\le
(1-\eta \mu)^{T_1}\left\|\theta_0-\theta^{\star}\right\|^2+\frac{\eta c}{\mu} \beta_{T_1}^2 .
\end{equation}

Choose $T_1$ so that $(1-\eta \mu)^{T_1}\left\|\theta_0-\theta^{\star}\right\|^2 \leq \varepsilon^2$. Then
\begin{equation}
\mathbb{E}\left\|\theta_{T_1}-\theta^{\star}\right\|
\le
\sqrt{\varepsilon^2+\frac{\eta c}{\mu} \beta_{T_1}^2}
\approx
\varepsilon+\sqrt{\frac{\eta c}{\mu}} \beta_{T_1}.
\end{equation}

For fixed $\beta_t=\beta_{\max}$ and the same timestep $T_1$, we have
\begin{equation}
\mathbb{E}\left\|\bar{\theta}_{T_1}-\theta^{\star}\right\|
\le
\varepsilon+\sqrt{\frac{\eta c}{\mu}} \beta_{\max }.
\end{equation}

Define
\begin{equation}
\delta_{\text{GPO}}^2=\varepsilon^2+\frac{\eta c}{\mu}\beta_{T_1}^2,\qquad
\delta_{\text{fixed}}^2=\varepsilon^2+\frac{\eta c}{\mu}\beta_{\max}^2,
\label{eq:defs_delta_Delta}
\end{equation}
so that $\delta_{\text{GPO}}^2<\delta_{\text{fixed}}^2$ for $\beta_{T_1}<\beta_{\max}$.

\paragraph{Local strong concavity of the return.}
Assume that $J$ is $\mu$-strongly concave in a basin $\mathcal B$ around the stationary point
$\theta^\star$ with $\nabla J(\theta^\star)=0$. Then for any $\theta\in\mathcal B$,
\begin{equation}
J(\theta^\star) - J(\theta)
\ge
\frac{\mu}{2}\,\|\theta-\theta^\star\|^2 .
\label{eq:strong_concavity_gap}
\end{equation}

Taking expectations yields
\begin{equation}
\mathbb{E}[J(\theta_{T_1})]
\ge
\mathbb{E}[J(\theta^\star)]
-\frac{\mu}{2}\,\mathbb{E}\|\theta_{T_1}-\theta^\star\|^2
\ \ge\
\mathbb{E}[J(\theta^\star)]-\frac{\mu}{2}\,\delta_{\text{GPO}}^2,
\end{equation}
and analogously for $\bar\theta_{T_1}$,
\begin{equation}
\mathbb E[J(\bar\theta_{T_1})]
\ge
\mathbb{E}[J(\theta^\star)]
-\frac{\mu}{2}\,\mathbb E\|\bar\theta_{T_1}-\theta^\star\|^2
\ \ge\
\mathbb{E}[J(\theta^\star)]-\frac{\mu}{2}\,\delta_{\text{fixed}}^2.
\label{eq:return_lower_max}
\end{equation}

Therefore, the fixed-range training yields a looser (smaller) lower bound on the expected return.
This shows that a smaller $\beta_t$ yields faster convergence to the local optimum in parameter space
and a tighter return guarantee after the same $T_1$ training steps.

\section{Late-Stage Behavior: Exploration and Asymptotic Performance}
\label{exploration_behavior}

The key idea is that a smaller action space induces lower gradient noise, which
leads to a smaller mean-squared parameter error and, consequently, a higher
expected accumulated return in a local quadratic regime.

Assume that the iterates remain in a basin $\mathcal B$ around a stationary point
$\theta^\star$ satisfying $\nabla J(\theta^\star)=0$.
Within this basin, the expected return admits a local quadratic form
\begin{equation}
J(\theta)
=
J(\theta^\star)
-\frac{1}{2}(\theta-\theta^\star)^\top H(\theta-\theta^\star),
\qquad
\mu I \preceq H \preceq L I,
\label{eq:local_quad_aligned}
\end{equation}
which is consistent with the $\mu$-strong concavity and $L$-smoothness assumptions
used in Appendix~\ref{app:Convergence_Error_Bound}.

Consider the stochastic policy-gradient update
\begin{equation}
\theta_{t+1}=\theta_t+\eta g_t,
\qquad
g_t=\nabla J(\theta_t)+\xi_t,
\end{equation}
where
\begin{equation}
\mathbb E[\xi_t\mid \theta_t]=0,
\qquad
\mathbb E[\|\xi_t\|^2\mid \theta_t]\le c\,\beta_t^2 .
\label{eq:noise_model_aligned}
\end{equation}
The fixed-range baseline follows
\[
\bar\theta_{t+1}=\bar\theta_t+\eta \bar g_t,
\qquad
\mathbb E[\|\bar\xi_t\|^2\mid \bar\theta_t]\le c\,\beta_{\max}^2 .
\]
Assume $\theta_0=\bar\theta_0$, $\beta_t<\beta_{\max}$ for all $t$, and
$0<\eta\le \mu/L^2$.

Define the parameter errors
\[
e_t:=\theta_t-\theta^\star,
\qquad
\bar e_t:=\bar\theta_t-\theta^\star.
\]
Under \eqref{eq:local_quad_aligned}, we have
$\nabla J(\theta_t)=-H e_t$, and thus
\[
e_{t+1}=(I-\eta H)e_t+\eta\xi_t,
\qquad
\bar e_{t+1}=(I-\eta H)\bar e_t+\eta\bar\xi_t.
\]
For $0<\eta\le \mu/L^2$, the matrix $(I-\eta H)$ is a contraction, and there exists
$\rho\in(0,1)$ such that
\[
\|I-\eta H\|_2^2 \le \rho .
\]
Taking conditional expectation and using the standard assumption that the cross
term vanishes yields
\begin{equation}
\mathbb E\|e_{t+1}\|^2
=
\rho\,\mathbb E\|e_t\|^2+\eta^2 c\,\beta_t^2,
\qquad
\mathbb E\|\bar e_{t+1}\|^2
=
\rho\,\mathbb E\|\bar e_t\|^2+\eta^2 c\,\beta_{\max}^2.
\label{eq:mse_recursions_aligned}
\end{equation}
%%%%%%%%%%%%%%%%%%%%%%%%%%%%%%%%%%%%%
\paragraph{Asymptotic ($t\to+\infty$) comparison.}
From \eqref{eq:mse_recursions_aligned}, taking $\limsup_{t\to\infty}$ on both sides yields
\begin{equation}
\limsup_{t\to\infty}\mathbb E\|e_{t+1}\|^2
\le
\rho\limsup_{t\to\infty}\mathbb E\|e_t\|^2+\eta^2 c\limsup_{t\to\infty}\beta_t^2.
\end{equation}
Let $\beta_\infty:=\limsup_{t\to\infty}\beta_t$. Since $\rho\in(0,1)$, rearranging gives the
steady-state bound
\begin{equation}
\limsup_{t\to\infty}\mathbb E\|e_t\|^2
\le
\frac{\eta^2 c}{1-\rho}\,\beta_\infty^2.
\label{eq:gpo_steady_mse}
\end{equation}
For the fixed-range baseline, $\beta_t\equiv \beta_{\max}$, hence
\begin{equation}
\limsup_{t\to\infty}\mathbb E\|\bar e_t\|^2
\le
\frac{\eta^2 c}{1-\rho}\,\beta_{\max}^2.
\label{eq:fixed_steady_mse}
\end{equation}
Therefore, since $\beta_\infty\le \beta_{\max}$,
\begin{equation}
\limsup_{t\to\infty}\mathbb E\|e_t\|^2
\le
\limsup_{t\to\infty}\mathbb E\|\bar e_t\|^2 .
\label{eq:steady_mse_compare}
\end{equation}

Using the quadratic form \eqref{eq:local_quad_aligned} and $H\preceq L I$, we have
$e_t^\top H e_t \le L\|e_t\|^2$ and thus
\begin{equation}
\mathbb E[J(\theta_t)]
=
J(\theta^\star)-\frac12\mathbb E[e_t^\top H e_t]
\ge
J(\theta^\star)-\frac{L}{2}\mathbb E\|e_t\|^2,
\label{eq:return_lb_gpo}
\end{equation}
and analogously
\begin{equation}
\mathbb E[J(\bar\theta_t)]
\ge
J(\theta^\star)-\frac{L}{2}\mathbb E\|\bar e_t\|^2 .
\label{eq:return_lb_fixed}
\end{equation}
Combining \eqref{eq:steady_mse_compare} with \eqref{eq:return_lb_gpo}--\eqref{eq:return_lb_fixed},
and using $\liminf x_t \le \limsup x_t$, yields the asymptotic (no-worse) dominance
\begin{equation}
\liminf_{t\to\infty}\mathbb E[J(\theta_t)]
\ \ge\
\liminf_{t\to\infty}\mathbb E[J(\bar\theta_t)].
\label{eq:asymp_return_compare}
\end{equation}

\section{Experiment}
\subsection{Tasks and Experimental Setup}
\label{app:implementation}

This part provides implementation details omitted from the main text.

\subsubsection{Observation}
\label{app:observation}

The observation vector is
\begin{equation}
o_t = [w_t,\, g_t,\, q,\, \dot{q},\, v_{\text{cmd}},\, \tau,\, \zeta_t]^{\top}.
\end{equation}

For the quadruped robot, $q$, $\dot{q}$, $\tau$, and $\zeta_t$ are 12-dimensional,
corresponding to the 12 actuated joints.
For the hexapod robot, these quantities are 18-dimensional, matching the 18 degrees
of freedom.

The command input $v_{\text{cmd}}$ is task-dependent.
For quadruped whole-body control, $v_{\text{cmd}}$ includes planar velocities
$(v_x, v_y)$, yaw rate $\omega_z$, a target base height, and a target pitch angle.
For hexapod locomotion control, $v_{\text{cmd}}$ consists of planar velocities
$(v_x, v_y)$ and yaw rate $\omega_z$ only.

\subsubsection{Estimator and Critic Conditioning}
\label{app:estimator}

The estimator processes a short observation history $o_{t-10:t}$ and outputs
\begin{equation}
e_t = [\hat{v}_t,\, z_t],
\end{equation}
where $\hat{v}_t$ is an estimate of the base linear velocity and $z_t$ is a latent
dynamics representation used for next-step observation prediction.

The actor conditions on $(o_t, e_t)$ and outputs joint torques.
The critic is conditioned only on the observation history $o_{t-10:t}$ to estimate
the state value.

\subsubsection{Fatigue Modeling}
\label{app:fatigue}

The fatigue-related state $\zeta_t$ follows \textsc{SATA}~\cite{li2025sata} and is
updated as
\begin{equation}
\zeta_t = \left(\zeta_{t-1} + |\tau| \cdot dt \right)\cdot \gamma ,
\end{equation}
where $dt$ is the control time step.
In all experiments, we set $\gamma = 0.95$.

\subsubsection{Reward Formulation}
\label{app:reward}

The reward function consists of task-specific tracking terms and shared
task-agnostic regularization terms.
Table~\ref{tab:rewards} lists all reward components and their corresponding weights.

\begin{table}[h!]
    \centering
    \renewcommand{\arraystretch}{1.3} % 增加行间距
    \setlength{\tabcolsep}{0pt} % 调整列间距
    \caption{Reward components for quadruped and hexapod robots and their corresponding weights (\(dt=0.005\))}
    \label{tab:rewards}
    \begin{tabular}{lcc}
    \toprule
    \textbf{Reward Terms} & \textbf{Equation} & \textbf{Weight} \\ 
    \midrule
    \multicolumn{3}{l}{\textbf{Quadruped Locomotion Objectives}} \\ 
    \(r_{\text{tracking},x}\) & \(\phi\left(v_x - v_x^{\text{cmd}}·\beta_t\right)\) & \(10dt\) \\ 
    \(r_{\text{tracking},y}\) & \(\phi\left(v_y - v_y^{\text{cmd}}·\beta_t\right)\) & \(5dt\) \\ 
    \(r_{\text{tracking},\text{yaw}}\) & \(\phi\left(\omega_{\text{yaw}} - \omega_{\text{yaw}}^{\text{cmd}}·\beta_t\right)\) & \(5dt\) \\ 
    \(r_{\text{tracking},\text{pitch}}\) & \(\phi\left(\theta - \theta^{\text{cmd}}·\beta_t\right)\) & \(5dt\) \\ 
    \(r_{\text{tracking},\text{height}}\) & \(\phi\left(h_b - h_b^{\text{cmd}}·\beta_t\right)\) & \(7dt\) \\ 
    \midrule
    \multicolumn{3}{l}{\textbf{Hexapod Locomotion Objectives}} \\ 
    \(r_{\text{tracking},x}\) & \(\phi\left(v_x - v_x^{\text{cmd}}·\beta_t\right)\) & \(10dt\) \\ 
    \(r_{\text{tracking},y}\) & \(\phi\left(v_y - v_y^{\text{cmd}}·\beta_t\right)\) & \(5dt\) \\ 
    \(r_{\text{tracking},\text{yaw}}\) & \(\phi\left(\omega_{\text{yaw}} - \omega_{\text{yaw}}^{\text{cmd}}·\beta_t\right)\) & \(5dt\) \\ 
   \(r_{\text{base height}}\) & \(\min(h_b, h_t)\) & \(7dt\) \\ 
    \midrule
    \multicolumn{3}{l}{\textbf{Auxiliary Posture Maintenance}} \\ 
    \(r_{\text{roll}}\) & \(|g_{\mathrm{y}}|\) & \(-5dt\) \\ 
    \(r_{\text{velocity},z}\) & \((v_z)^2\) & \(-5dt\) \\ 
    \(r_{\text{joint limits}}\) & $\sum \left[ (q_{\min} - q)^+ + (q - q_{\max})^+ \right]$ & \(-5dt\) \\ 
    \(r_{\text{fatigue}}\) & \(\zeta \cdot |\tau_{d} \cdot \kappa_{\text{scale}}|\) & \(-0.05dt\) \\ 
    \(r_{\text{joint acceleration}}\) & \(\ddot{q}^2\) & \(-1e-6dt\) \\ 
    \bottomrule
    \end{tabular}
    \end{table}

% Appendixes should appear before the acknowledgment.

% \section*{ACKNOWLEDGMENT}

% The preferred spelling of the word ÒacknowledgmentÓ in America is without an ÒeÓ after the ÒgÓ. Avoid the stilted expression, ÒOne of us (R. B. G.) thanks . . .Ó  Instead, try ÒR. B. G. thanksÓ. Put sponsor acknowledgments in the unnumbered footnote on the first page.

%%%%%%%%%%%%%%%%%%%%%%%%%%%%%%%%%%%%%%%%%%%%%%%%%%%%%%%%%%%%%%%%%%%%%%%%%%%%%%%%

\end{document}